\documentclass[letterpaper, 10 pt, conference]{ieeeconf}  

\IEEEoverridecommandlockouts                              

\usepackage{graphicx} 
\usepackage{amsmath} 
\usepackage{amssymb}  
\usepackage{booktabs}

\usepackage{algorithm}
\usepackage{algpseudocode}
\usepackage{graphicx}
\usepackage{xcolor}

\usepackage{amsthm}
\usepackage{hyperref}

\newcommand{\ie}{\textit{i.e.}}
\newcommand{\eg}{\textit{e.g.}}

\title{\LARGE \bf
Human Grasp Classification for Reactive Human-to-Robot Handovers
}

\usepackage[font=scriptsize,labelfont=bf]{caption}
\author{Wei Yang*$^1$, Chris Paxton*$^1$, Maya Cakmak$^{1,2}$, and Dieter Fox$^{1,2}$
\thanks{* Equal Contribution}
\thanks{$^{1}$ NVIDIA, USA
        {\tt\small \{weiy,cpaxton,dieterf\}@nvidia.com}}%
\thanks{$^{2}$ University of Washington, USA
        {\tt\small mcakmak@cs.washington.edu}}%
}

\begin{document}

\maketitle
\thispagestyle{empty}
\pagestyle{empty}

\begin{abstract}
Transfer of objects between humans and robots is a critical capability for collaborative robots.
Although there has been a recent surge of interest in human-robot handovers, most prior research focus on robot-to-human handovers.
Further, work on the equally critical human-to-robot handovers often assumes humans can place the object in the robot's gripper.
In this paper, we propose an approach for human-to-robot handovers in which the robot meets the human halfway, by classifying the human's grasp of the object and quickly planning a trajectory accordingly to take the object from the human's hand according to their intent.
To do this, we collect a human grasp dataset which covers typical ways of holding objects with various hand shapes and poses, and learn a deep model on this dataset to classify the hand grasps into one of these categories.
We present a planning and execution approach that takes the object from the human hand 
according to the detected grasp and hand position,
and replans as necessary when the handover is interrupted.
Through a systematic evaluation, we demonstrate that our system results in more fluent handovers versus two baselines. We also present findings from a user study ($N=9$) demonstrating the effectiveness and usability of our approach with naive users in different scenarios. 
More results and videos can be found at \url{http://wyang.me/handovers}.
\end{abstract}

\section{INTRODUCTION}\label{section:intro}
Giving and taking objects to and from humans are fundamental capabilities for collaborative robots across applications from manufacturing to physical assistance in the home. 
A growing community of researchers in robotics have been studying the problem of enabling fluent human-robot handovers. 
Most work focuses on transfer of objects from the robot to the human, assuming the human can just place the object in the robot's gripper for the reverse.
This approach is not feasible in scenarios where the human needs to pay attention to their task at hand, such as performing a surgery, or where the human has limited mobility and arm movement due to an impairment.
Such scenarios require more reactive handovers that can adapt to the way that the human is presenting the object to the robot and meet them half way to take the object.


One of the key challenges in making human-to-robot handovers reactive is reliable and continuous perception of the object and the human.
One strategy is to estimate the human hand pose as well as the 6D object pose by borrowing off-the-shelf methods from the computer vision community. 
However, state-of-the-art methods for hand pose estimation~\cite{ge2018point,iqbal2018hand,ge20193d} and object pose estimation~\cite{xiang2017posecnn} focus on only the hand or the objects independently.
Although few recent methods jointly estimate hand and object poses while the hand is interacting with the object~\cite{hasson2019learning, zimmermann2019freihand,hampali2019honnotate}, their accuracy is limited when the object and the hand are occluded by each other.

\begin{figure}[bt]
    \centering
    \includegraphics[width=0.49\columnwidth]{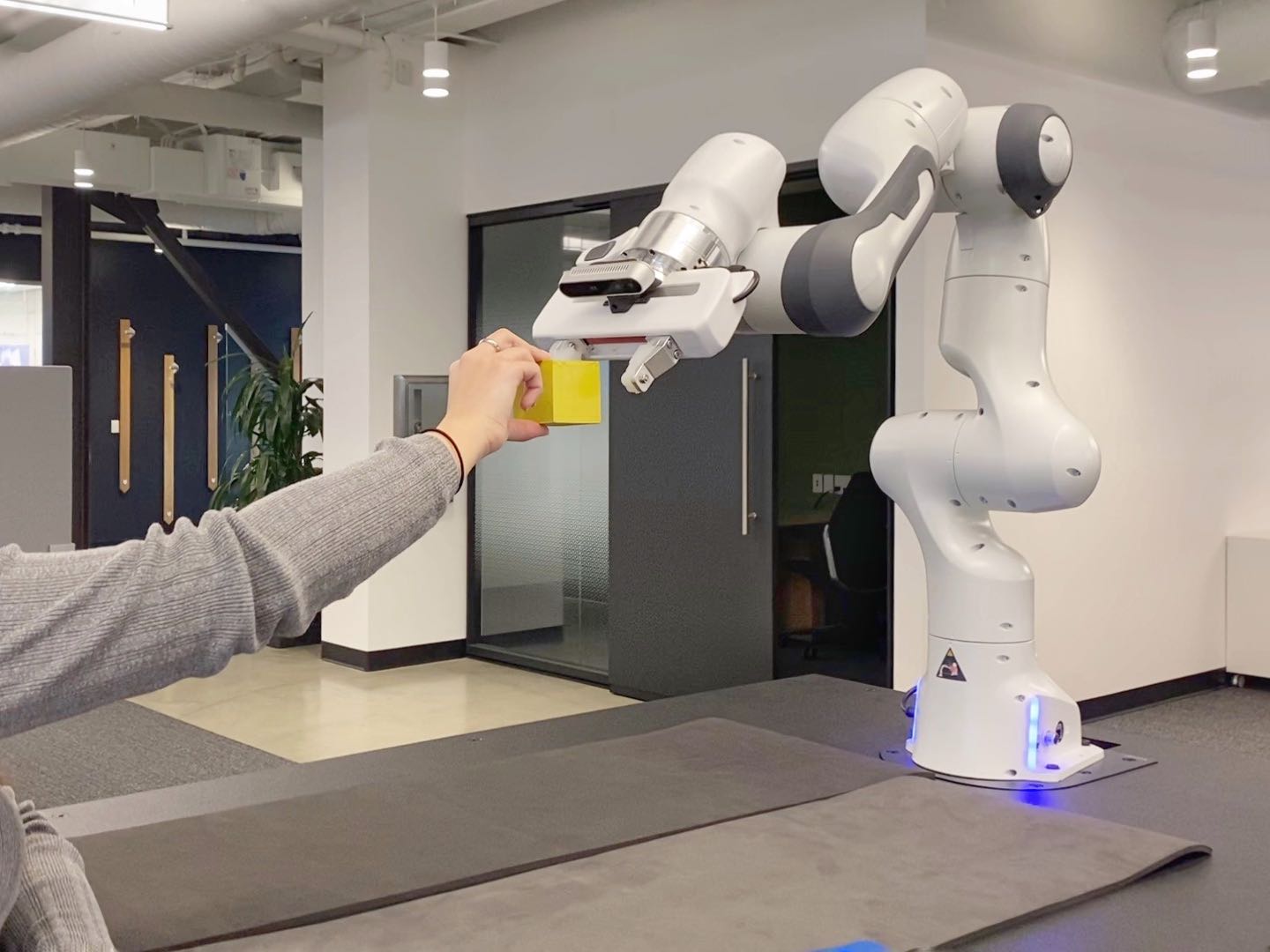}
    \includegraphics[width=0.49\columnwidth]{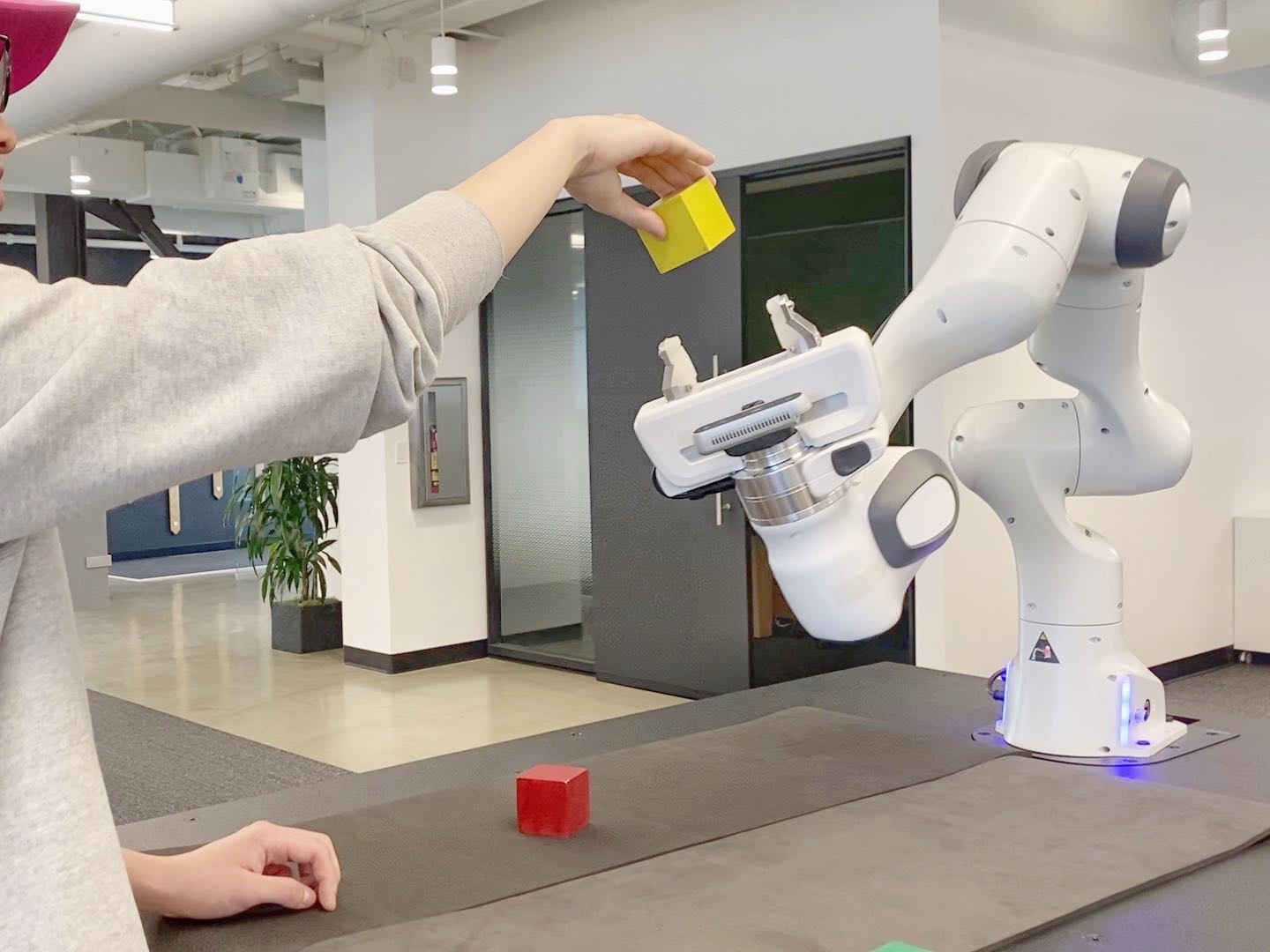}
    \vskip 0.15cm
    \includegraphics[width=0.49\columnwidth]{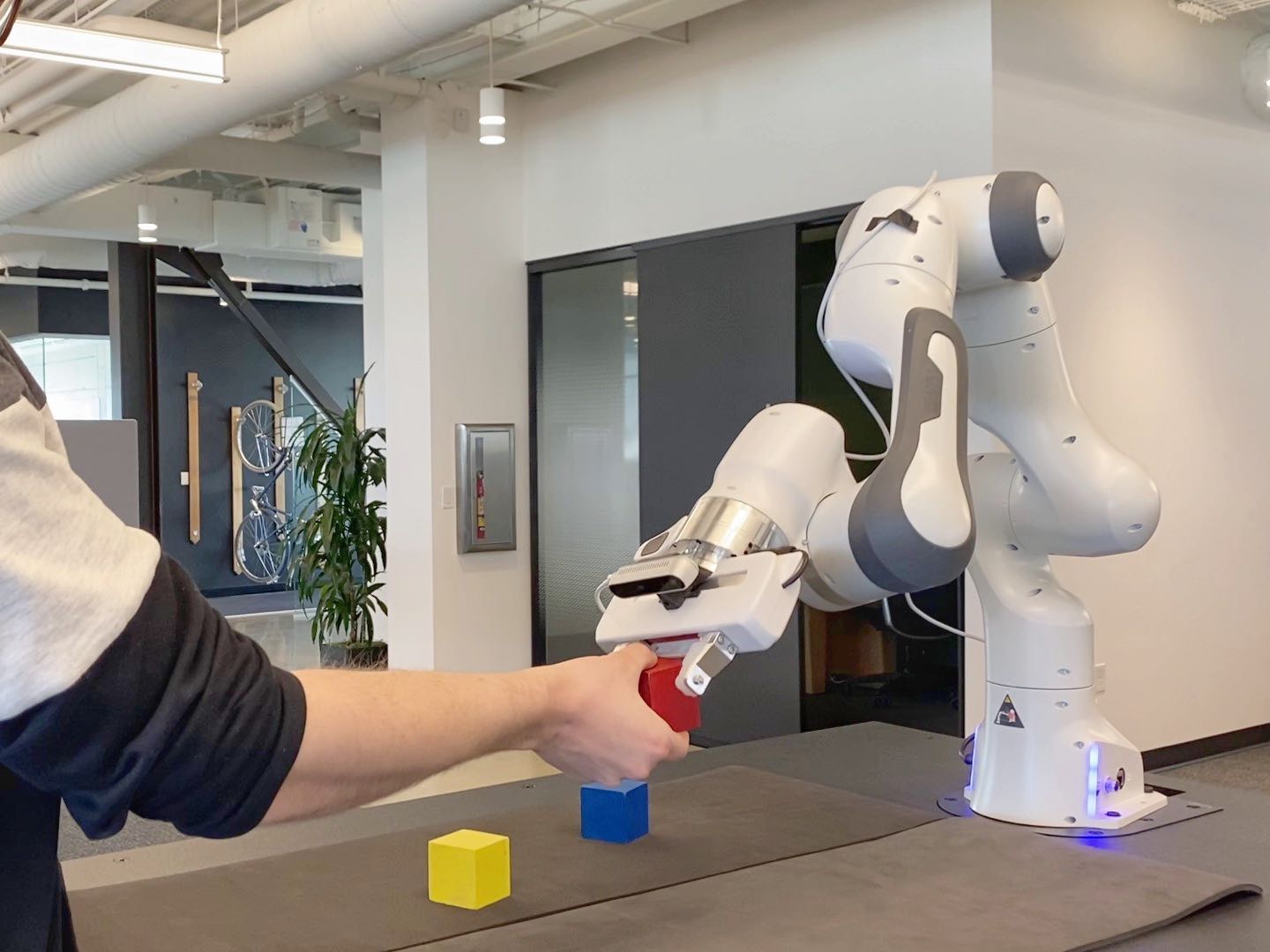}
    \includegraphics[width=0.49\columnwidth]{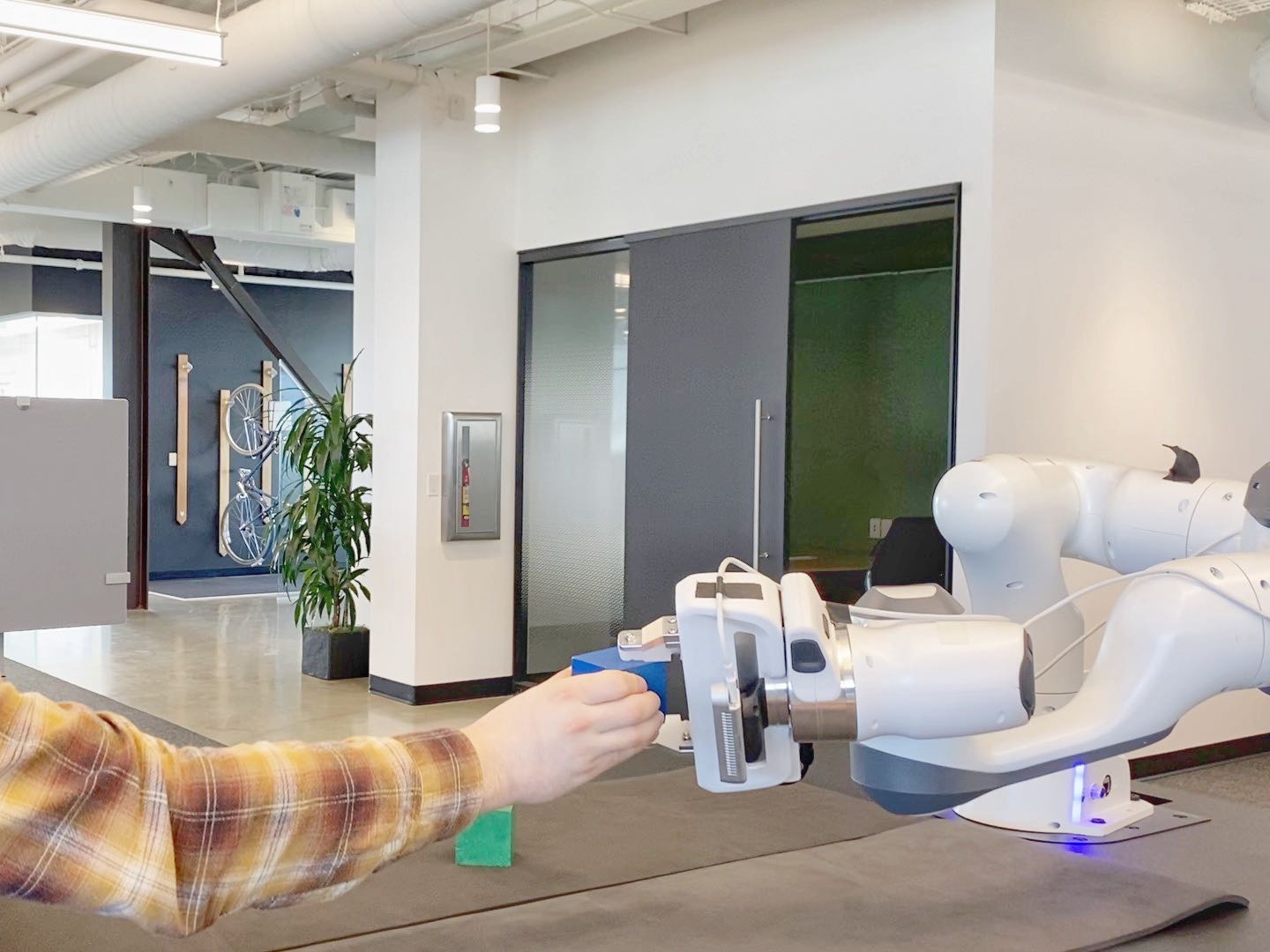}
    \caption{Humans hand objects over in different ways. They can present the object on their palm or use a pinch grasp and present the object in different orientations. Our system can determine which grasp a human is using and adapt accordingly, enabling a reactive human-robot handover.}
    \label{fig:grasps-cover}
    \vskip -0.5cm
\end{figure}




In this paper we propose to address the problem of perception for human-to-robot handovers by formulating it as a hand grasp classification problem. 
Specifically, we discretize the ways in which humans can hold small objects into several categories (Fig.~\ref{fig:grasps-cover}) and we collect a dataset to learn a deep model that classifies a given human hand holding an object into one of those grasp categories.
We model the handover task as a Robust Logical-Dynamical System following on previous work~\cite{paxton2019representing}, which generates motion plans that avoid contact between the gripper and the human hand given the human grasp classification.
We compare our system with two baseline methods, one without inferring the human hand pose and the other relying on independent hand and object pose estimation, demonstrating higher success rate and time efficiency of our approach over the two baselines. 
We also present a user study ($N=9$) demonstrating the effectiveness of our approach with naive users, both while they are attentive to the robot and while they are focused on a secondary task. Participants agreed that our system is collaborative, trustworthy and aware of the humans' actions.

The main contributions of this paper are: 
(1) hand-object interaction reasoning for handovers posed as a classification problem, via a dataset that covers a wide range of hand shapes and poses; 
(2) a system that adaptively plans robot grasps for taking the object from the human, so that the robot can respond to the human fluidly and naturally; and
(3) experimental results demonstrating improvements over the baseline methods, and a user study validating our approach with naive users.


\section{RELATED WORK}\label{section:related_work}
Human-robot handovers have recently become a popular topic within human-robot collaboration \cite{bauer2008human} across a multitude of application areas from collaborative manufacturing \cite{koene2014relative,unhelkar2014comparative,wang2018controlling} to assistance in the home \cite{mainprice2012sharing,Huang2015AdaptiveCS,cakmak2011using,grigore2013joint}.
A large majority of this work focuses on robot-to-human handovers in which the robot starts with an object in hand and transfers it to the human. 
A key challenge is choosing parameters of the robot's actions to optimize for a fluent handover.
This includes the choice of object pose and robot's grasp on the object, taking into account user comfort \cite{aleotti2012comfortable}, preferences based on subjective feedback \cite{cakmak2011human}, affordances and intended use of the objects after the handover \cite{aleotti2014affordance,chan2014determining,bestick2016implicitly,chan2019affordance,cini2019choice}, motion constraints of the human \cite{mainprice2012sharing}, social role of the human \cite{kato2019handovers}, and configuration of the object when being grasped before the handover \cite{ardon2018towards}. 
Other work emphasizes parameters of the trajectory to reach the handover pose, exploring the approach angle \cite{unhelkar2014comparative}, starting pose of trajectory in contrast to the handover pose \cite{cakmak2011using}, motion smoothness \cite{de2016neural}, object release time \cite{han2019effects}, estimated human wrist pose \cite{maeda2017probabilistic,sidiropoulos2019human}, relative timing of handover phases \cite{kshirsagar2019specifying}, and ergonomic preferences of humans \cite{parastegari2017modeling}.
While some work focuses on offline computation of handover parameters, most recent work involves perception of the human to enable reactive handovers \cite{peternel2017towards,maeda2017probabilistic,kupcsik2018learning,zhou2018early}.



A number of user studies have been conducted to validate different handover approaches and provide empirical evidence, such as people's preference among alternative ways of handing objects \cite{cakmak2011using,cakmak2011human}, impact of robot behaviors such as gaze \cite{admoni2014deliberate,moon2014meet}, or difference between novice and experienced users \cite{zu2017hand}.
Some work has explored human-human handovers to characterize movement properties \cite{becchio2010toward,huber2008human,shibata1997analysis}, grip force patterns \cite{chan2012grip}, use of social cues \cite{shi2013model}, or failure recovery strategies \cite{parastegari2018failure}.


Although less frequent, some work has explored human-to-robot handovers, \ie, how robots may take objects from humans \cite{edsinger2007human, aleotti2012comfortable}.
Pan et al. explored the problem of detecting handover intent by the human based on skeleton tracking data obtained in human-human handovers \cite{pan2017automated}.
Other work enabled human-to-robot handovers via wearable sensing on the human \cite{wang2018controlling}.
Vogt et al. proposed to learn a controller to both give and receive objects from a single demonstration of handovers between two humans \cite{vogt2018one}.
Most closely related to our research, Marturi et al. investigated grasping of moving objects to enable human-to-robot handovers \cite{marturi2019dynamic}.


Given our focus on perception of humans to enable reactive handovers, prior work on human hand  pose estimation are also highly relevant. 
Though 3D human hand pose estimation is being actively studied in computer vision, most of the existing work focuses on monocular RGB images~\cite{iqbal2018hand,ge20193d}, which results in insufficient 3D localization for handovers. 
Some~\cite{handa2019dexpilot,ge2018point,handa2019dexpilot} use depth information for more precise hand pose estimation. However, they are mostly trained on data with a bare hand only due to the difficulty to collect data of hands interacting with objects, and are tend to fail in circumstances that the object is with close proximity with the hand. 
Instead of understanding hand pose and object pose in isolation, some recent work estimates hand-object manipulations~\cite{zimmermann2019freihand,hasson2019learning,hampali2019honnotate}. 
While promising, these methods either trained on synthetic data~\cite{hasson2019learning} which requires to bridge the sim-to-real gap, or on sensor data within a close range which is not suitable for distant hand recognition for handovers~\cite{zimmermann2019freihand, hampali2019honnotate}.


Our system for task execution is based on Robust Logical-Dynamical Systems~\cite{paxton2019representing}, an approach for automatically creating reactive task plans for robots. The idea is to constantly identify the present logical state and reactively replan to handle uncertainty and changes in logical state, an approach that's been proven useful for dealing with partially-observable environments (\eg,~\cite{garrett2019online}). For our purposes, the task model can be thought of in a similar way to Behavior Trees~\cite{colledanchise2018behavior}, a method for representing complex tasks that has previously been shown useful for human-robot collaboration~\cite{paxton2017costar,paxton2018evaluating}.

\section{HUMAN GRASP CLASSIFICATION}\label{section:hand_model}
\begin{figure*}[t]
  \includegraphics[width=1\linewidth]{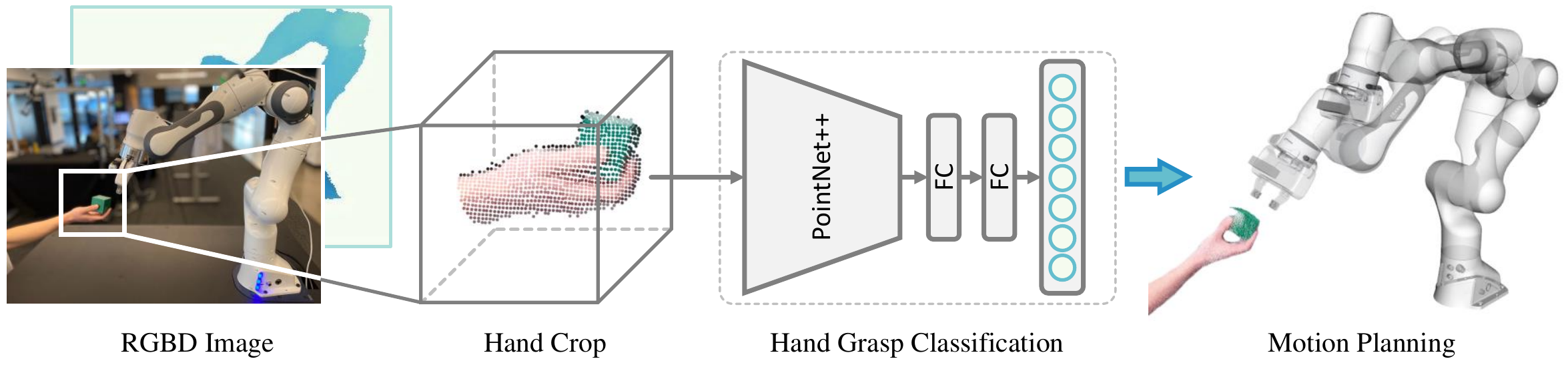}
  \caption{An overview of our handover framework. 
  The framework takes the point cloud centered around the hand detection, and then uses a model inspired by PointNet++~\cite{qi2017pointnet++} to classify it as one of seven grasp types which cover various ways objects tend to be grasped by the human user. 
  Out task model will then plan the robot grasps adaptively.}
  \label{fig:overview}
  \vskip -0.5cm
\end{figure*}

\begin{figure*}[bt]
  \includegraphics[width=1\linewidth]{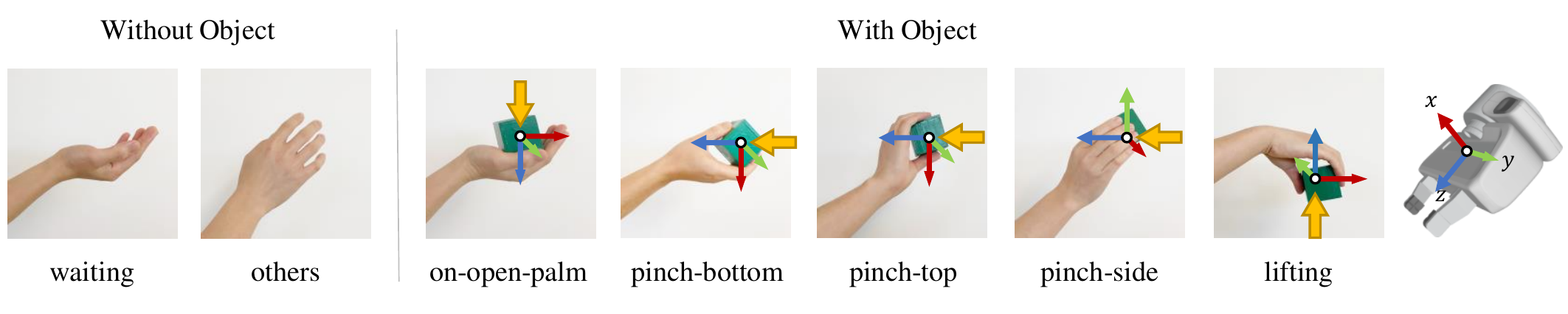}
  \caption{Five human grasp types with two empty hand types which cover various ways objects tend to be grasped by the human user. 
  These are associated with different robot canonical grasp directions in order to minimize human's efforts during handovers (illustrated by the coordinate system and the yellow arrow). Best viewed in color.}
  \label{fig:grasp_types}
  \vskip -0.5cm
\end{figure*}

When the robot takes an object from humans, the motion should be adjusted according to the way that the object is grasped by the human hand. 
Otherwise the robot could behave in an nonintuitive way or even grasp human fingers. 
As illustrated in Fig.~\ref{fig:overview}, our proposed handover framework addresses this issue by taking the point cloud centered around the human hand detected by the Azure Body Tracking SDK~\cite{k4a}, and then estimating the hand grasp class based on how the block is grasped by the human hand. Our task model will then adaptively plan robot grasps. 

In this section, we first define a discrete set of human grasps which describe the way that the object is grasped by the human hand for the task of handover. 
Then we present how we train a deep neural network to predict the human grasp categories based on the point cloud. 
Finally, we discuss how we adjust the orientation of robot grasps according to the human grasps.

\subsection{Human Grasp Definition}
Inspired by the study on the human grasp taxonomy~\cite{feix2015grasp}, we discretize the common human grasps for the task of human-robot handover into seven categories, as shown in Fig.~\ref{fig:grasp_types}: If the hand is grasping a block, then the hand pose can be categorized as \textit{on-open-palm, pinch-bottom, pinch-top, pinch-side}, or \textit{lifting}. If the hand is not holding anything, it could be either \textit{waiting} for the robot to handover an object or just doing nothing specific (\textit{others}).

\subsection{Human Grasp Dataset}
In order to learn a model to classify the human grasps, we create a dataset which covers eight subjects with various hand shape and hand pose by using an Azure Kinect RGBD camera. 
Specifically, we show an example image of a hand grasp to the subject, and record the subject performing similar poses from twenty to sixty seconds. 
The whole sequence of images are therefore labeled as the corresponding human grasp category. 
During the recording, the subject can move his/her body and hand to different position to diversify the camera viewpoints. We record both left and right hands for each subject. In total, our dataset consists of $151,551$ images. 

\subsection{Human Grasp Classification Model}
Instead of learning deep features with ConvNets on depth images, we adopt the recently developed PointNet++~\cite{qi2017pointnet++} on point clouds for human grasp classification due to its efficiency and its success on many robotics applications such as markerless teleoperation system~\cite{handa2019dexpilot} and grasps generation~\cite{murali2020clutteredgrasping}. 
Our backbone network consists of four set-abstraction layers to learn point features and a three-layer perceptron with batch normalization, ReLu and Dropout for global feature learning and human grasp classification.
Given a point cloud cropped around the hand, the network classifies it into one of the defined grasp categories, which would be used for further robot grasp planning.

\subsection{Canonical Robot Grasp Directions}
We associate each human grasp type with a canonical robot grasp direction in order to minimize human's effort during the human-to-robot handovers. 
As shown in Fig.~\ref{fig:grasp_types}, the coordinates denote the canonical robot grasp frames in the camera frame. 
The motivation is to reduce the chance for the robot to grab human's hand while keep its motion and trajectory as natural and smooth as possible. 



\section{TASK MODEL}\label{section:task_model}

Our task model is based on Robust Logical-Dynamical Systems~\cite{paxton2019representing}. This
represents tasks as a list of reactively-executed operators $o$ with certain properties. Each operator is a tuple $o = \{L_P, L_R, L_E, \pi\}$,
where $L_P$ is a set of logical preconditions on entering $o$, $L_R$ is a set of run conditions that must hold while execution of $o$ is ongoing, and $L_E$ is the set of logical effects that will be true. The operator is also associated with a policy $\pi$ which will generate the necessary controls to achieve effects $L_E$. The policy and predicates can be learned from data~\cite{kase2020transferable}, but in our case they are specified manually.
Given a plan, we choose the highest-priority operator whose preconditions are met, checking conditions at 10 hz so we can quickly respond to changes.

\begin{figure*}[bt]
\centering
\begin{tabular}{c c c c}
\toprule
     (1) Wait for Human & (2) Find Plan, Follow Plan & (3) Grasp Object & (4) Drop Object \\
\midrule
     \includegraphics[width=0.46\columnwidth]{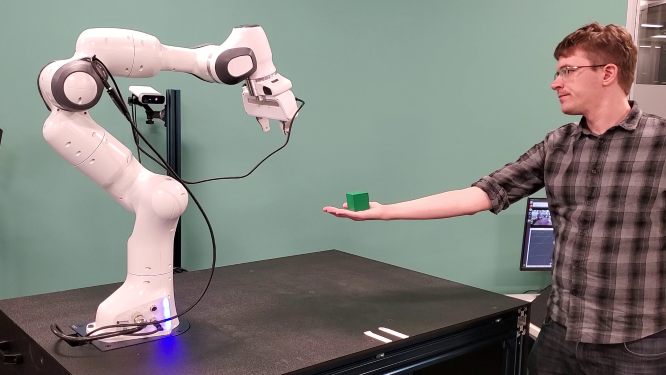} & 
     \includegraphics[width=0.46\columnwidth]{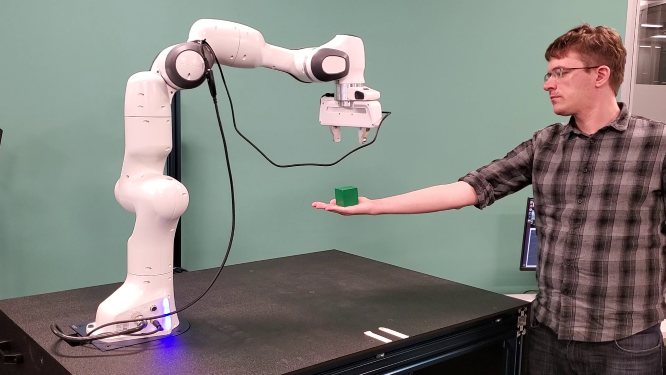} & 
     \includegraphics[width=0.46\columnwidth]{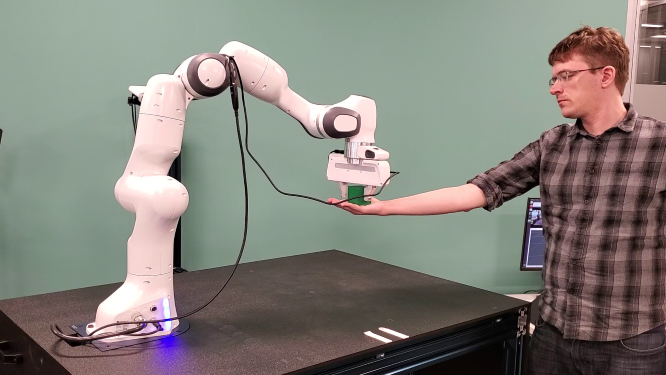} & 
     \includegraphics[width=0.46\columnwidth]{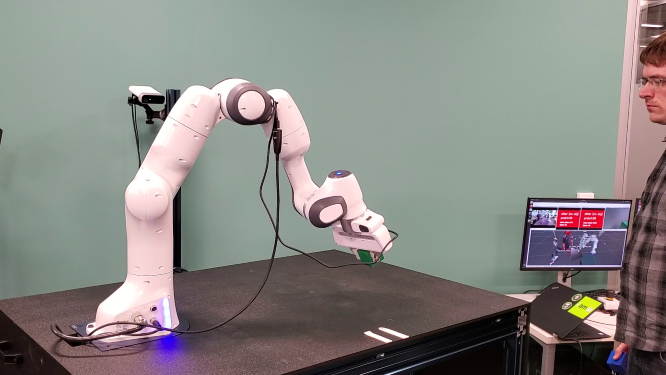} \\
     \includegraphics[width=0.46\columnwidth]{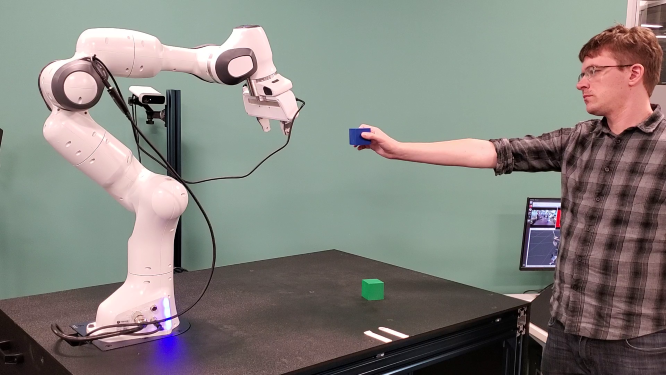} & 
     \includegraphics[width=0.46\columnwidth]{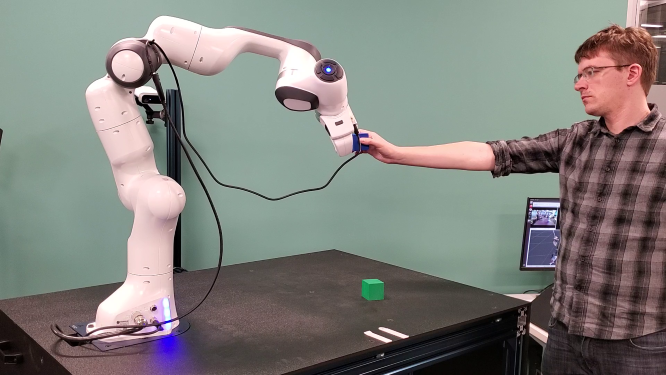} & 
     \includegraphics[width=0.46\columnwidth]{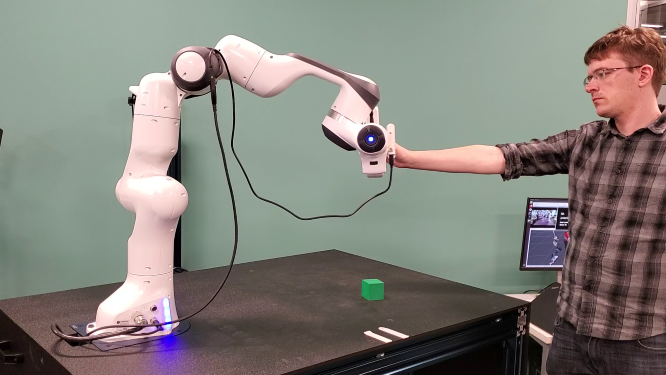} & 
     \includegraphics[width=0.46\columnwidth]{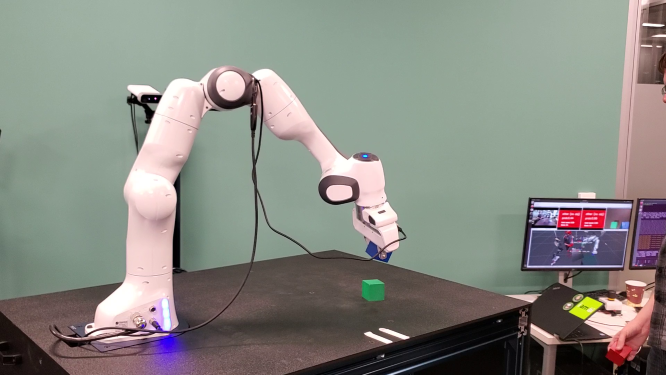} \\
\bottomrule
\end{tabular}
\caption{Examples of task plan execution with the given system. The important operators are (1) Waiting at the ``home'' position for the human to enter the workspace and for position estimates to stabilize, (2) choosing a safe grasp plan, (3) moving to the grasp position and taking the block from the human, and (4) dropping the object on the table. These steps can be interrupted by the human. Descriptions of policies and preconditions are in Sec.~\ref{section:task_model}.}
\label{fig:task-plan}
\vskip -0.5cm
\end{figure*}

Fig.~\ref{fig:task-plan} gives an overview of the different steps in our final task plan. The system has to adapt to different possible grasps, reactively choosing the correct way to approach the human user and take the object from them. Until it gets a stable estimate of how the human wants to present the block, it stays in a ``home'' position and waits.

Instead of just using reactive local planning, we found we needed to plan and make intelligent decisions based on a large number of possible grasps in order to find the one that would be the most natural to the human user, as discussed in prior work~\cite{paxton2017costar,paxton2018evaluating}. We describe the extra predicates and operators involved in this reactive planning process below.
Table~\ref{tab:operators} shows the task plan, in order of descending priority.


\begin{table}[bt]
    \centering
    \begin{tabular}{l p{0.6\linewidth}}
    \toprule
    Operator & Preconditions \\
    \midrule
         Open gripper &  $\texttt{has\_obj} \wedge \texttt{gripper\_fully\_closed}$ \\
         Wait for human & $\neg (\texttt{stable} \wedge \texttt{hand\_over\_table} \wedge \texttt{hand\_has\_obj})$ \\
         Avoid human & $\texttt{too\_close\_to\_hand} \wedge \neg \texttt{in\_approach\_region}$ \\
         Drop object & $\texttt{at\_drop\_position} \wedge \texttt{has\_obj}$ \\
         Go to drop & $\texttt{has\_obj}$ \\
         Grasp object & $\texttt{in\_approach\_region} \wedge \texttt{has\_goal} \wedge \texttt{is\_goal\_valid}$ \\ 
         Follow plan & $\texttt{has\_goal} \wedge \texttt{is\_goal\_valid}$ \\
         Find plan & $\texttt{has\_feasible\_goals}$ \\
         Find feasible goals &  $\texttt{stable} \wedge \texttt{hand\_over\_table} \wedge \texttt{hand\_has\_obj}$ \\
        \bottomrule
    \end{tabular}
    \caption{Operators and corresponding preconditions $L_P$ for task execution and reactive execution. Operators are listed in descending order of priority; if all the preconditions are true, we execute the associated operator regardless of what the previously executed operator was.}
    \label{tab:operators}
\vskip -0.5cm
\end{table}

\textbf{Wait for human.}
We compute several predicates determining how the robot should interact with the hand: \texttt{stable}, \texttt{hand\_over\_table}, and \texttt{hand\_has\_obj}, and \texttt{too\_close\_to\_hand}.
The \texttt{hand\_over\_table} predicate corresponds to whether or not these observations are in a specified volume over the table depicted in Fig.~\ref{fig:task-plan}, and \texttt{stable()} is true if the hand is not moving and the hand has been observed for at least 5 timesteps (0.5 seconds). We defined this based on the velocity:
\[
    \texttt{stable()} = \| x_{t-1} - x_{t} \|_2 < \lambda,
\]
\noindent with position $x$ and time $t$, for a threshold $\lambda$. The robot will wait at the home position if these conditions are not true.

\textbf{Avoid human.} If \texttt{too\_close\_to\_hand()} is true for either hand and the robot is not in the approach region corresponding to a particular grasp, the robot will attempt to avoid the hand and will move back to the home positions. We define \texttt{too\_close\_to\_hand()} to be true if the Euclidean distance between the end effector and the hand is less than 20 cm.

\textbf{Find feasible goals.} In order to ensure that the robot's motions are safe, instead of the purely reactive policies used in prior work~\cite{paxton2019representing}, we plan whole trajectories for execution. If \texttt{stable()}, \texttt{hand\_over\_table()}, and \texttt{hand\_has\_obj()}, then the robot will attempt to take the object from the hand, using the canonical grasp pose shown in Fig.~\ref{fig:grasp_types}.

In order to find a valid trajectory $\xi$, the robot must first find a valid grasp pose, so we add the \texttt{has\_goal} and \texttt{is\_goal\_valid} predicates. If either of these is false, we search for a reasonable goal pose.

The planner will create a list of goal pose candidates and associated standoff positions. There are ten options, at rotations of $\theta_y \in \{ -\pi/4, -\pi/8, 0, \pi/8, \pi/4\}$ around the $y$-axis in Fig.~\ref{fig:grasp_types}, and $\theta_z \in \{ 0, \pi \}$ around the $z$-axis. Any feasible grasp poses (\ie, grasp poses with a corresponding inverse kinematics solution). Both grasp and standoff position must be collision-free and have a valid IK solution in order to be considered feasible goal options. We also add a constraint that the robot should never occlude its view of the object when determining if states are valid.

\textbf{Find plan.} If the planner has a list of goal options, it will then sort them according to their distance from the current joint configuration and attempt to find a motion plan to the standoff position using RRT-Connect~\cite{kuffner2000rrt}. If the system can both find a grasp pose and a motion plan, the robot will execute a sub-policy to follow this motion plan.

However, a human might move their hand or change how they are holding an object in their hand. A goal is only considered valid (as per the \texttt{is\_goal\_valid} predicate) if it has an associated motion plan, and if the object has not moved within some threshold of where it was first observed. If the object moves too much, the robot must stop, and the task model will instantly transition back to finding a new grasp.

\textbf{Grasp Object.} Once a motion plan has completed, the robot should be at a standoff pose and have an associated goal pose -- the expected position of the object in the human hand. These two poses define an \textit{approach region} -- a conical volume within which the robot can move to approach the object, as described in our prior work~\cite{paxton2019representing}. 
Once the gripper closes, if the robot is at its goal pose, the \texttt{has\_obj} predicate is set to true. The grasp operator may occlude the object, so we execute this as our only blocking, open-loop action.

\textbf{Open Gripper.} If \texttt{has\_obj} is true, indicating that the robot believes it is holding an object, we may still be wrong because the object moved or the pose estimate was off. We add a \texttt{gripper\_fully\_closed} predicate, saying that the gripper closed all the way. If both conditions are true, we set $\texttt{has\_obj}$ to false, and the robot will revert to a different state.

\textbf{Move to drop} and \texttt{Drop object.} The drop position is a single joint-space position; our robot will find a safe, collision-free motion plan. If it is at the drop position, it will open the gripper and put the object on the table.

\section{SYSTEMATIC EVALUATION}\label{section:exp}
In this section, we perform a systematic evaluation, where we compare our method to several baseline approaches with multiple metrics. We also report the performance of our human grasp classification model.

\subsection{Experimental Setup}

We performed a systematic of the entire system, including the classification model described in Sec.~\ref{section:hand_model} and the task model described in Sec.~\ref{ssec:metrics}, on a range of different hand positions and the grasps shown in Fig.~\ref{fig:grasp_types}.
We used two different Franka Panda robots, mounted on identical tables in different locations, as shown in Fig.~\ref{fig:task-plan}. A human user handed four colored blocks over to the robot, one at a time. 
During the systematic evaluation, we tested each of three approaches for determining which grasp pose to use for taking an object from the human:






\textbf{Simple Baseline}: 
waits until it sees the block in a human hand and takes it from the hand, using a fixed grasp orientation. The human hand is detected via an off-the-shelf system; in our case, the Microsoft Azure body tracker~\cite{k4a}.
    
\textbf{Hand Pose Estimation}: A state estimation-based version of the system, in which we use the human hand pose from the Azure body tracker to infer grasp direction. 
    
\textbf{Ours}: The proposed system, classifying human grasps based on depth information as described in Sec.~\ref{section:hand_model}.

All variants executed the same task model, as described in Sec.~\ref{section:task_model}. The order in which these three test cases were provided was randomized. 
Users used their right hand to present blocks to the robot. 

\subsection{Evaluation Metrics} \label{ssec:metrics}

We evaluate the system performance with a set of metrics computed during trials. These were computed automatically and logged while users were performing the task.

\textbf{Planning Success Rate:}
the number of times the \texttt{follow\_plan} operator was able to execute successfully, bringing the robot to its standoff pose, and measures certainty both of the human and the system. 

\textbf{Grasp Success Rate:} how often the robot was able to successfully take the object from the human, versus total number of grasp attempts that it made.

\textbf{Action execution Time:} tracks the amount of time it took to execute a single planned trajectory, grasp a block, and place it on the table. This is higher if the robot must take a longer path to grasp the block from the human.

\textbf{Total Execution Time:} the amount of time it took to execute all planned paths, including replanning because the human moved or because of the changing of the way of grasp.

\textbf{Trial Duration:} Time since human hand was first detected until the trial was complete.


\subsection{Results}

\begin{table*}[bt]
    \centering
    \caption{Results for handover performance on our quantitative metrics. Planning success rate indicates how often the system needed to replan its approach, versus grasp success rate as the number of times the system successfully took the object.}
    \label{tab:hand-tracking}
    \begin{tabular}{l c c c c c}
        \toprule
         & Planning Success Rate & Grasp Success Rate & Action Execution Time (s) & Total Execution Time (s) & Trial Duration (s) \\
         \midrule
         Simple Baseline & 42.1\% & 66.7\% & 11.37 & 20.93 & 21.59 \\
         Hand Pose Estimation & 29.6\% & 80.0\% & 15.10 & 36.34 & 36.46 \\
         Ours & 64.3\% & 100\% & 13.20 & 17.34 & 18.31 \\
         \bottomrule
    \end{tabular}
\end{table*}

Table~\ref{tab:hand-tracking} shows results on each of our main metrics during the systematic evaluation.
Our method consistently improves the success rate and reduces the total execution time and the trial duration compared with the other two baseline methods, which proves the efficacy and the reliability of our method. 

The only exception is the \textit{Action Execution Time}, where the \textit{simple baseline} is often faster. This is because the simple baseline does not plan as adaptively as the others; it would not try to attempt an unusual grasp. This means that time from a successful approach to dropping the object is, on average, notably lower.

\begin{figure}
    \centering
    \includegraphics[width=1\linewidth]{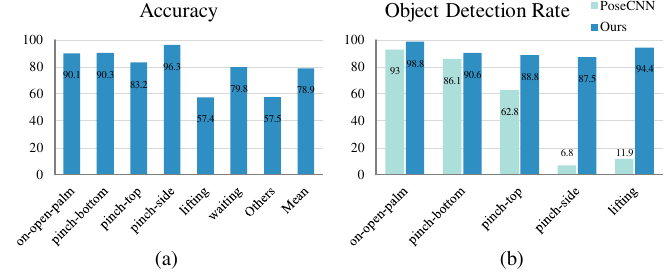}
    \caption{
    (a) The accuracy of the human hand grasp classification. 
    (b) The comparison of the object miss-detection rate between our hand states classification and PoseCNN~\cite{xiang2017posecnn}. In many cases the hand occludes the object, meaning that it is very difficult to get an accurate pose estimate. 
    }
    \vspace{-1em}
    \label{fig:acc_detection_rate}
\end{figure}

\noindent\textbf{Evaluation of Human Grasp Classification}: 
We evaluate our hand grasp classification model on a validation set collected with a subject which is unseen during the training procedure. The classification accuracy is reported in Fig~\ref{fig:acc_detection_rate} (a), which demonstrates the good generalization ability of our model on unseen subjects.

In addition, we conduct an experiment to evaluate the detection rate, \ie, whether there is an object in the hand, to give us an insight on how robust the handover system is against the occlusion. 
We compare the detection rate of our hand grasp classification model (with/without object) to that of a state-of-the-art object detection method~\cite{xiang2017posecnn}. 
The result is reported in Fig.~\ref{fig:acc_detection_rate}(b). 
We can see that our human grasp model achieves higher detection rate and is more robust compared with~\cite{xiang2017posecnn} especially when heavy occlusion occurs (\eg, $87.5\%$ vs. $6.8\%$ for  \textit{pinch-side} and $94.4\%$ vs. $11.9\%$ for \textit{lifting}).





\section{USER STUDY}\label{section:user}


We also performed a user study in order to determine if our system allowed for fluid human-robot collaboration. Wee recruited nine users, ages $20$ to $36$. Of these, two were female and seven were male. The average age was $30.44 \pm 4.74$ years. 
The study consisted of three rounds:

\textbf{Freeform}: first, users were given four blocks and instructed to stand in front of the table and hand the blocks over to the robot one at a time. They were instructed that the robot would only take blocks if their hand is still, but they could hold the blocks any way they liked.

\textbf{Attentive}: 
Next we demonstrated the set of five human grasps 
shown in Fig.~\ref{fig:grasp_types}: \textit{pinch-top, pinch-bottom, pinch-side, lifting}, and \textit{on-open-palm}. We then told the participants to hand over four blocks again. 
We encouraged them to try the predefined hand grasps, but they were able to use any others.
    
\textbf{Distracted}: Finally, we tested user performance in the presence of a distraction. Users watched a music video on YouTube \footnote{\url{https://youtu.be/4jd6dNrJRh4}} and counted the number of faces that appeared, while handing over all four blocks to the robot.


In addition to the metrics described in Sec.~\ref{ssec:metrics}, we also
counted the following statistics during the user study: (a) number of times robot gripper contacted human fingers, (b) number of times users changed the grasp they were using, and (c) number of times they changed their hand position.
%
After each trial, participants were asked to describe any problems they experienced while handing the blocks to the robot.
After all three trials were done, we asked them to filled a Likert scale questionnaire and explain their answers.


\subsection{Results}

\begin{figure}
    \centering
    \includegraphics[width=\columnwidth]{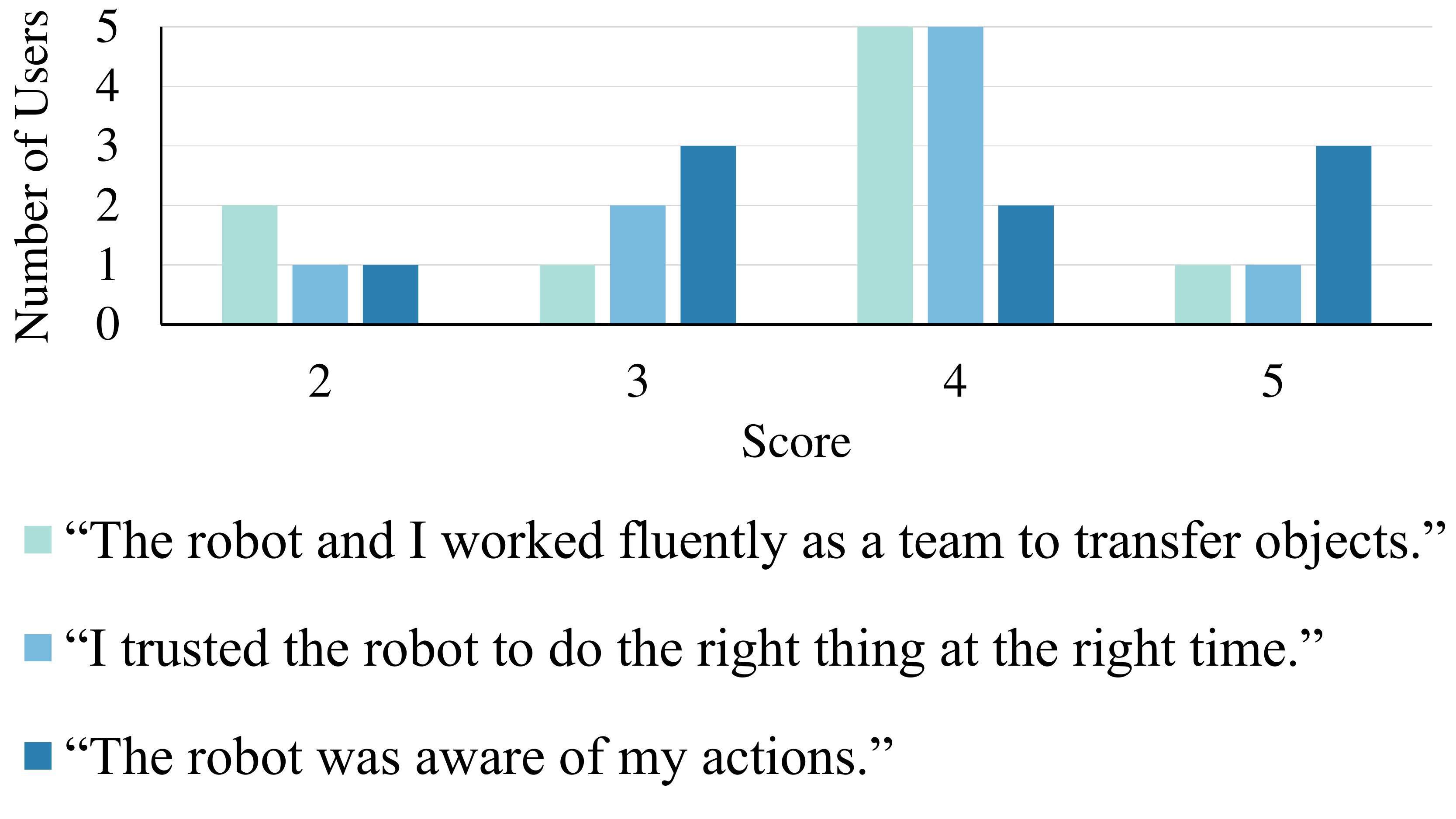}
    \caption{Results form the Likert scale questionaire given to users during the usability study. Users thought they worked together fluently with the robot, although there were several issues. Users were asked questions at the end of the study.}
    \label{fig:likert_scale}
\end{figure}

Fig.~\ref{fig:likert_scale} shows the results of the questionnaire given to our study participants. There was a range of responses, but users said that they worked fluently with the robot and trusted it to do the right thing, although they noted several common issues when asked for feedback. They also believed that the robot was aware of their actions.

\begin{table*}[bt]
    \centering
    \caption{Quantitative results from the user study. Users were able to complete tasks quickly even when they were distracted and had to concentrate on a different scenario.}
    \label{tab:study}
    \begin{tabular}{l c c c c c}
        \toprule
         & Planning Success Rate & Grasp Success Rate & Action Execution Time (s) & Total Execution Time (s) & Trial Duration (s) \\
         \midrule
         Freeform & 32.7\% & 67.3\% & 13.21 & 25.99 & 26.92 \\
         Attentive & 40.0\% & 90.0\% & 14.85 & 23.84 & 24.75 \\
         Distracted & 29.8\% & 67.9\% & 11.08 & 26.08 & 27.02 \\
         \midrule
         Overall & 33.6\% & 73.6\% & 13.05 & 25.31 & 26.24 \\
         \bottomrule
    \end{tabular}
\end{table*}

We also computed our quantitative metrics on user data, as seen in Table~\ref{tab:study}. Approaches and grasps were less successful when users were distracted, but times are similar.
Users counted an average of $12.88 \pm 3.48$ faces in the music video, when the correct number was $13$. This implies many of them felt certain level of confident of the handover system and were paying a good amount of attention to the video.

\subsection{Discussion}

\begin{figure}[bt]
    \centering
    \includegraphics[width=0.21\textwidth]{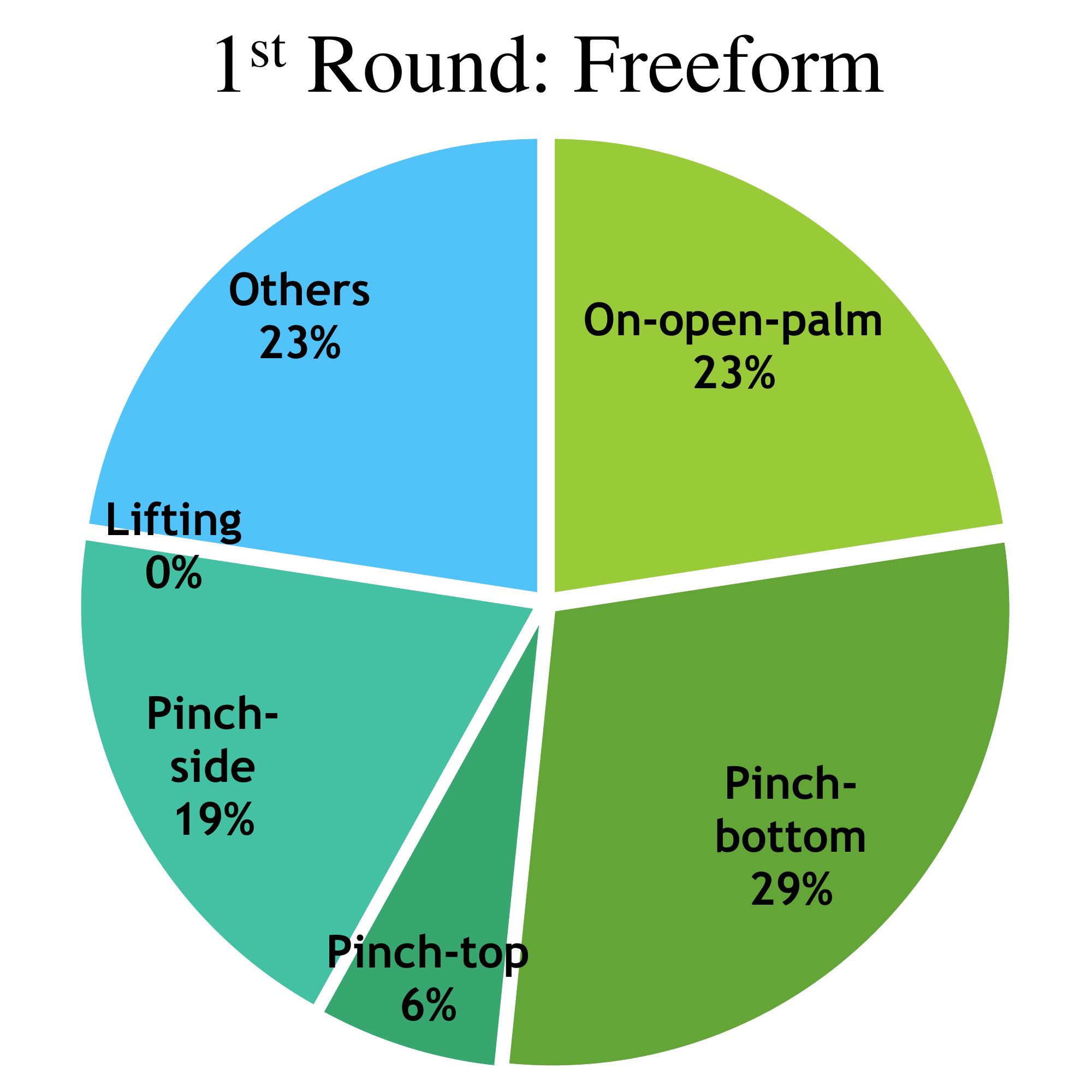}
    \includegraphics[width=0.21\textwidth]{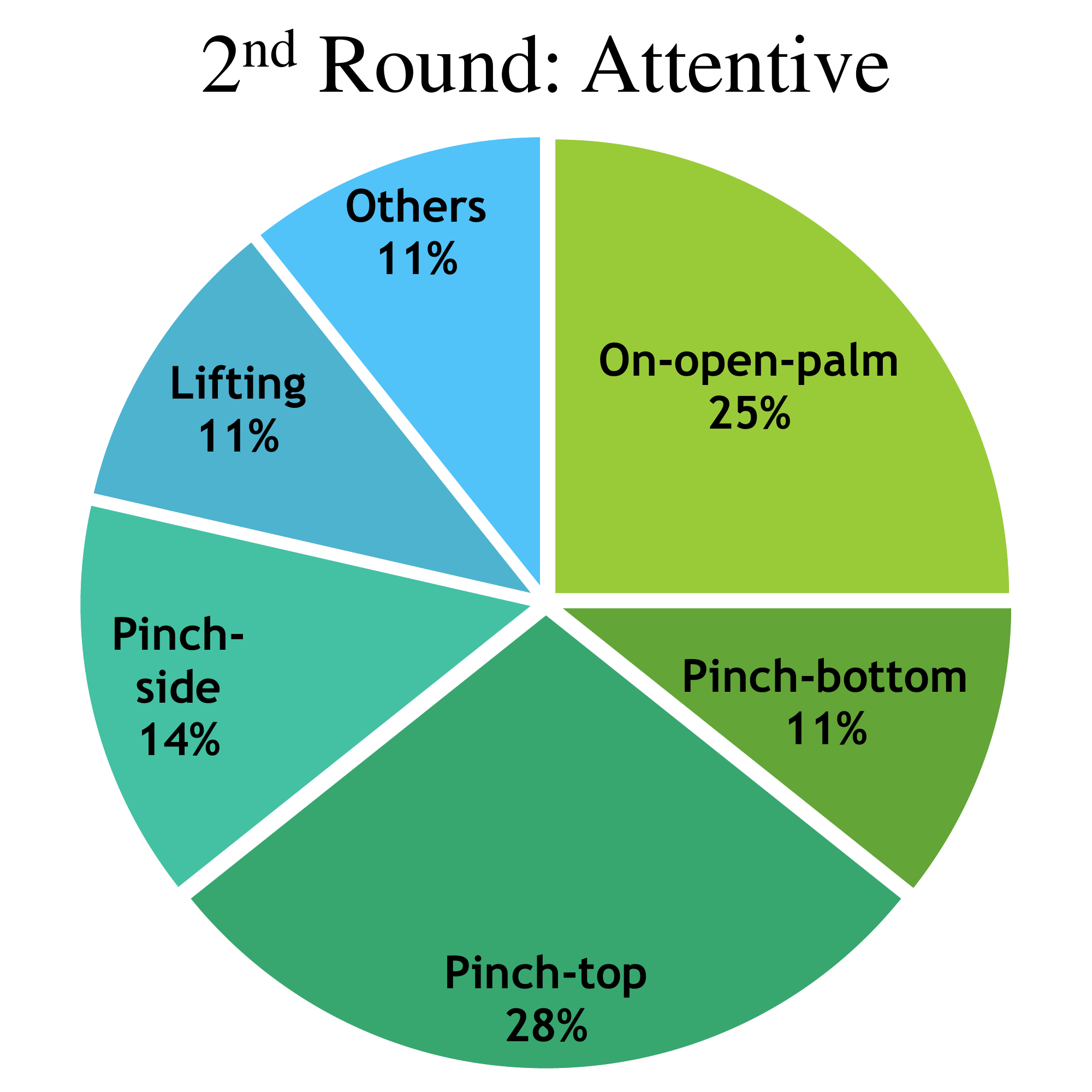}
    \includegraphics[width=0.21\textwidth]{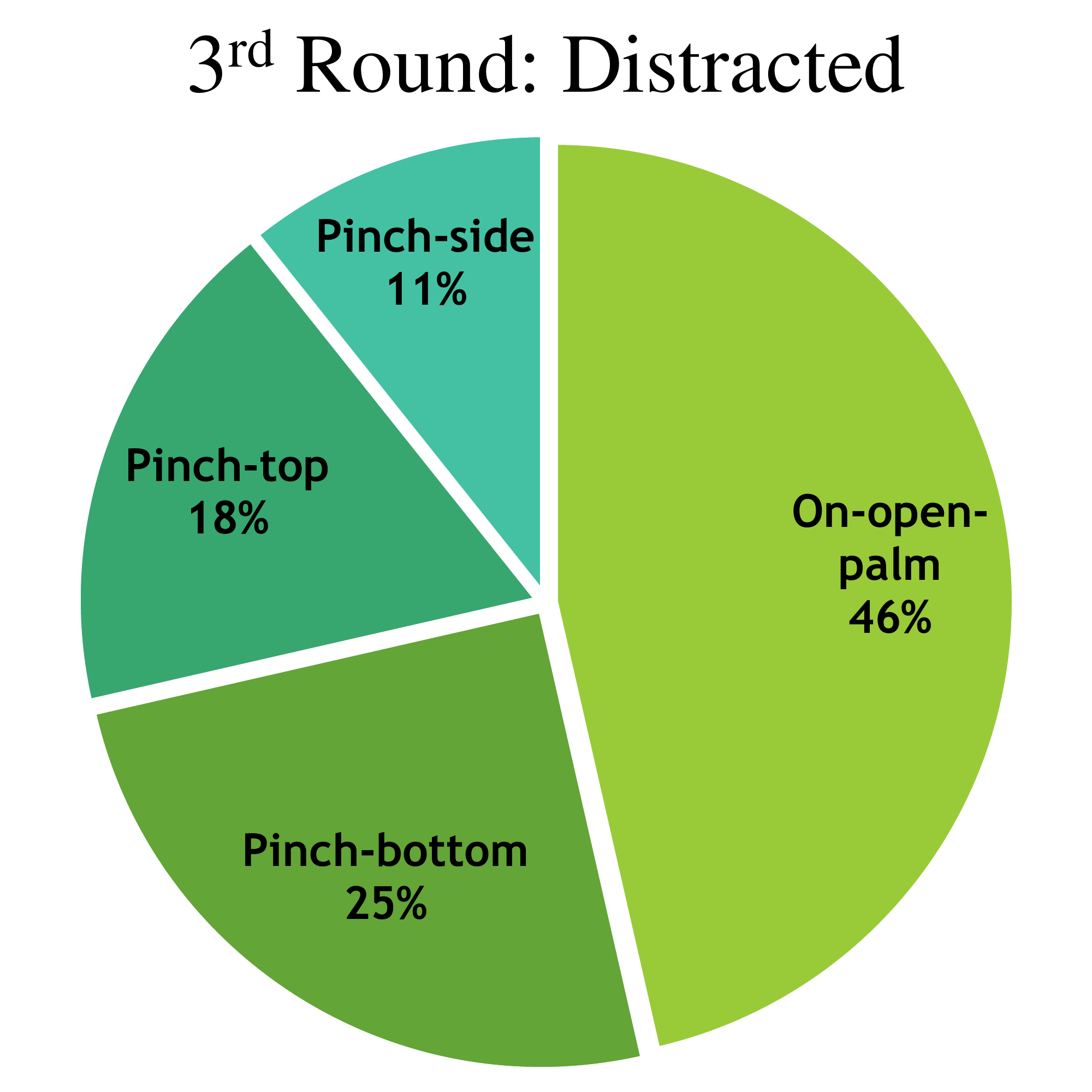}
    \includegraphics[width=0.21\textwidth]{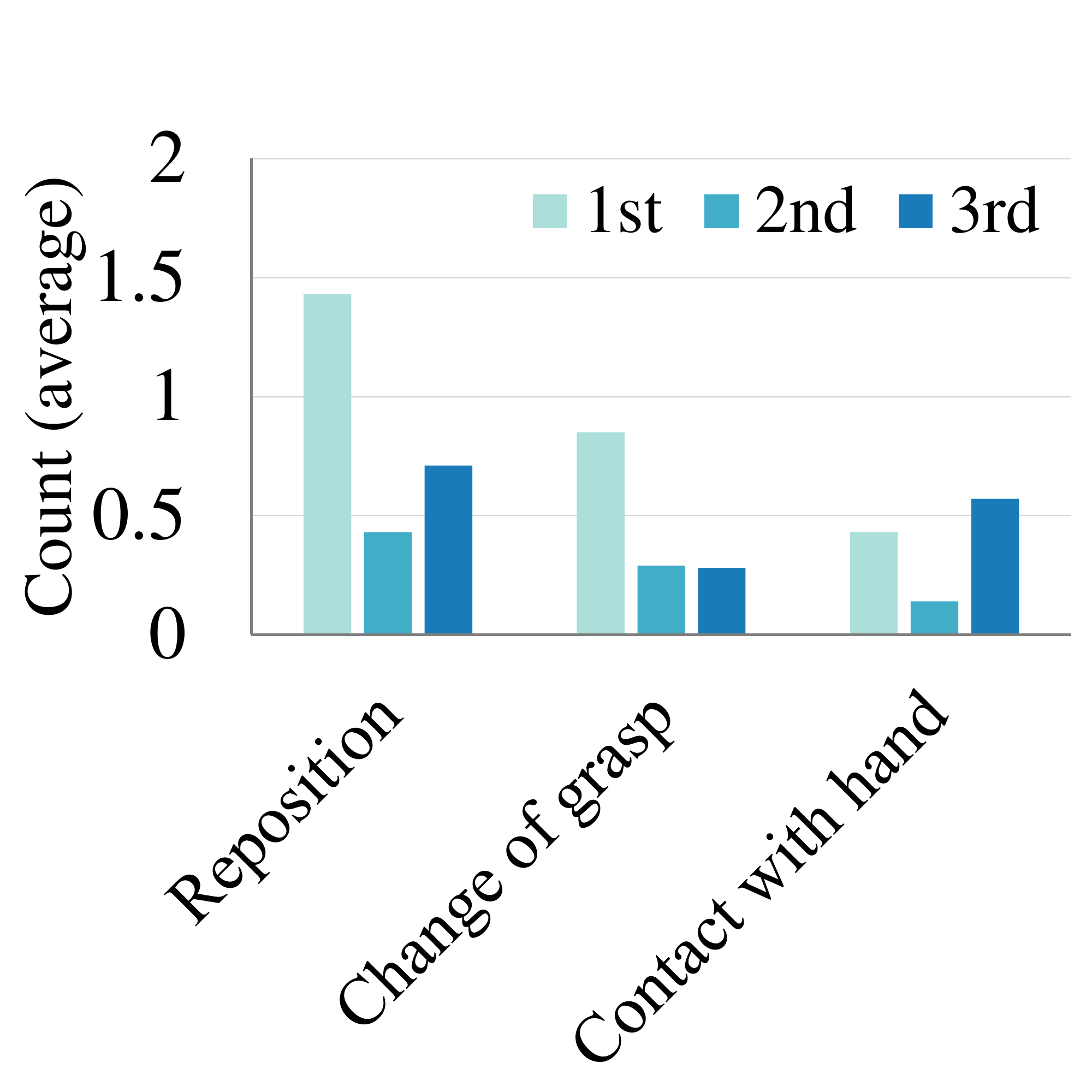}
    \caption{
    Coverage of human grasps during the usability study. \textit{Others} denotes the user grasps that are not included in our human grasp dataset. 
    Our human grasp dataset covers $77\%$ of the user grasps even  before the users were informed the ways of grasp included in our system (see \textit{Freeform}). We also report the average times that the users repositioned their hand, changed the way of grasp, and were contacted by the gripper. 
    }
    \label{fig:user_study_statistics}
\end{figure}

\begin{figure}[bt]
	\centering
	\small
    \includegraphics[width=0.24\linewidth]{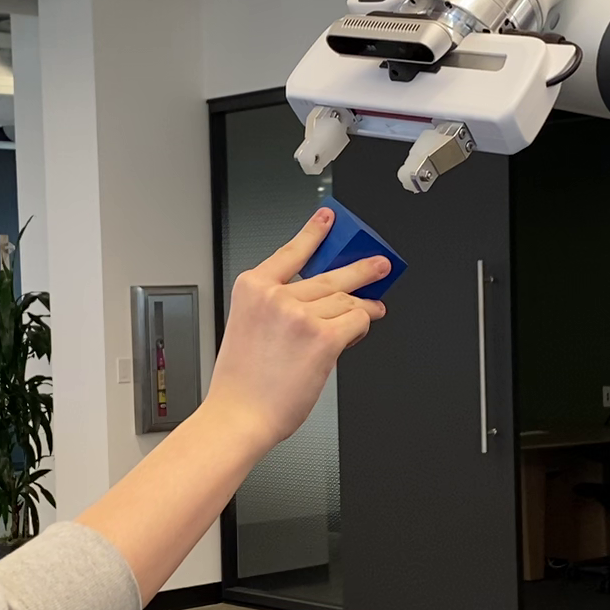}
	\includegraphics[width=0.24\linewidth]{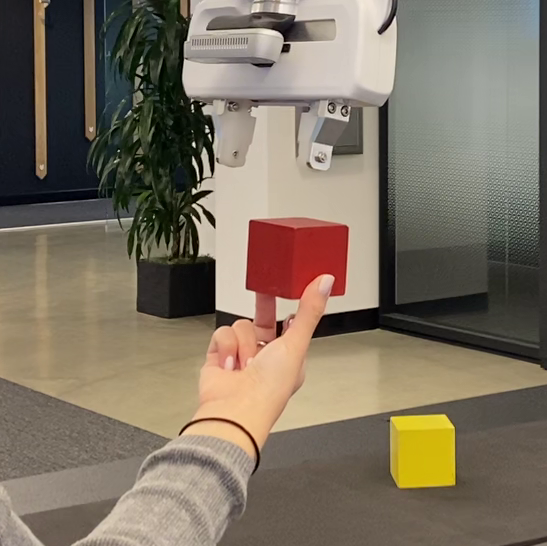}
	\includegraphics[width=0.24\linewidth]{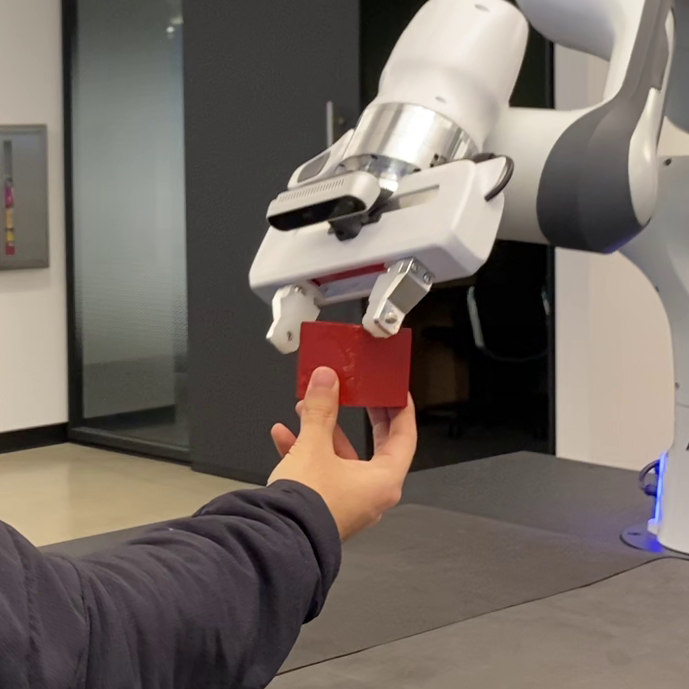}
	\includegraphics[width=0.24\linewidth]{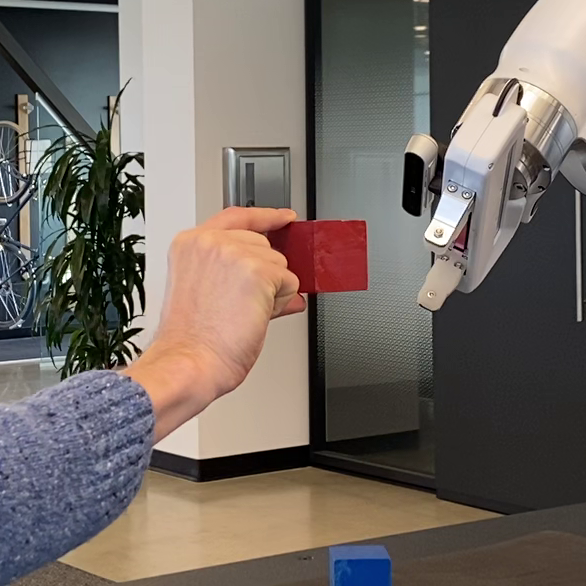} 
	
	\caption{Outliers grasps which do not appear in our training dataset, and are examples of the types of grasps where our system showed higher uncertainty, leading to slightly worse handover performance.}
	\label{fig:outliers}
\end{figure}

First, we report the coverage of different ways of grasp during the usability study in Fig.~\ref{fig:user_study_statistics}. 
In general, our definition of human grasps covers $77\%$ of the user grasps even before they know the ways of grasps defined in our system (see \textit{Freeform} in Fig.~\ref{fig:user_study_statistics}, which proves a good coverage of our human grasp design. 
While our system can deal most of the unseen human grasps, they tend to lead to higher uncertainty and sometimes would cause the robot to backoff and replan.
Some of these unseen grasps are shown in Fig.~\ref{fig:outliers}. This suggests directions for future research; ideally we would be able to handle a wider range of grasps that a human might want to use.

We also report the average number of times that users repositioned their hand, changed their way of grasp, and were contacted by the gripper in Fig.~\ref{fig:user_study_statistics}. 
During the final, distracted test, users had to reposition or change their ways of grasp more often compared with the first two rounds. 
Several complained about their fingers being pinched, or saw the robot fail to grasp objects. One specifically ``chose to use the palm-facing-up hand pose'' to minimize risk of failure; another ``had to look at the robot every 10s or so.'' 

This issue came up for several reasons. One of the most important is that many of the grasp poses are very hard for the robot to reach -- they are at the edges of its configuration space, and may require some complex motions to get there. 
In the future, we should strive to make the whole system more legible, indicating which blocks the robot wants to move to and how it wants to get there~\cite{dragan2013legibility}. 

In general, our users quickly noticed that the robot was trying to grab the block in an unobtrusive way as possible. 
They also noted a slight inaccuracy during the robot's grasps and approaches, but this doesn't seem to have been a major issue. One said about the robot ``it may require my assistance in slightly moving towards where it expected the goal to be.'' In other words, even though the robot's grasp pose might be slightly off due to an inaccurate hand pose or occluded object, the robot and human together were reliably able to execute the handoff.

From a usability perspective, our trajectories weren't always totally legible. The underlying motion planner used in Sec.~\ref{section:task_model} was based on RRT-connect~\cite{kuffner2000rrt}, and sometimes it would make surprising choices to reach grasp positions. For example, one user said after the first experiment, ``There was one of the trajectories that had a small detour.'' Another said ``Holding the block in such a way that the robot needed to rotate its grasp 90 degrees seemed to cause problems.'' This is because the robot would often be unclear about what the position and orientation of the hand was, and would end up being uncertain. In the end, the motion plans could be hard to interpret -- ``I was still unsure of what the actual behavior will be like.''

Users did notice that after the second round of experiments, when they were shown how to grasp the objects, that the system was more reliable and easier to work with. One users said ``the estimate of the object/hand position seemed more precise'' after they were taught these grasps. 

During the final, distracted test, users were more nervous. Several complained about their fingers being pinched, or saw the robot fail to grasp objects. One specifically ``chose to use the palm-facing-up hand pose'' to minimize risk of failure; another ``had to look at the robot every 10s or so.'' 

This issue came up for several reasons. First, many of the grasp poses are very hard for the robot to reach -- they are at the edges of its configuration space, and may require some complex motions to get there. 
Second, the users could also reposition their hands unintentionally while being distracted. 
In the future, we should strive to make the whole system more legible, indicating which blocks the robot wants to move to and how it wants to get there~\cite{dragan2013legibility}. 


\section{CONCLUSIONS}\label{section:conclusion}
We described a system for enabling fluid human-robot handovers via classifying different types of grasp. In the future, we will make the planning system more flexible and  support more grasp types. We believe the same approach could also be applied to many other types of human-robot collaboration.
The main limitation of our approach is that it applies only to a single set of grasp types, so additionally we plan to learn the correct grasp poses for different grasp types from data instead of using manually-specified rules. Based on user feedback, we also plan to make robot motions more legible and friendly.







\bibliographystyle{IEEEtran}
\bibliography{IEEEexample}

\end{document}